\documentclass[conference,letterpaper]{IEEEtran}
\IEEEoverridecommandlockouts
\usepackage{cite}
\usepackage{amsmath,amssymb,amsfonts}
\usepackage{bm}
\usepackage{algorithmic}
\usepackage{graphicx}
\usepackage{tikz}
\usepackage[caption=false,font=footnotesize]{subfig}
\usepackage[letterpaper,left=0.72in,right=0.72in,top=0.75in,bottom=1in]{geometry}
\usepackage{hyperref}
\usepackage{tabularx}
\usepackage{booktabs}
\usepackage{textcomp}
\usepackage{xcolor}
\usepackage{multirow}
\def\BibTeX{{\rm B\kern-.05em{\sc i\kern-.025em b}\kern-.08em
    T\kern-.1667em\lower.7ex\hbox{E}\kern-.125emX}}
\begin{document}
\newcommand\submittedtext{
  \footnotesize This work has been submitted to the IEEE for possible publication. Copyright may be transferred without notice, after which this version may no longer be accessible.}

\newcommand\submittednotice{
\begin{tikzpicture}[remember picture,overlay]
\node[anchor=south,yshift=10pt] at (current page.south) {\fbox{\parbox{\dimexpr0.65\textwidth-\fboxsep-\fboxrule\relax}{\submittedtext}}};
\end{tikzpicture}
}
\title{\vspace{1em} \LARGE \bf X-TRACK: Physics-Aware xLSTM for Realistic Vehicle Trajectory Prediction
\author{\IEEEauthorblockN{Aanchal Rajesh Chugh}
\IEEEauthorblockA{\textit{Technische Hochschule Augsburg} \\
\textit{TTZ Landsberg am Lech}\\
Augsburg, Germany \\
aanchal.rajesh.chugh@tha.de}
\and
\IEEEauthorblockN{Marion Neumeier}
\IEEEauthorblockA{\textit{Technische Hochschule Ingolstadt} \\
\textit{AImotion}\\
Ingolstadt, Germany \\
marion.neumeier@thi.de}
\and
\IEEEauthorblockN{Sebastian Dorn}
\IEEEauthorblockA{\textit{Technische Hochschule Augsburg} \\
\textit{TTZ Landsberg am Lech}\\
Augsburg, Germany \\
sebastian.dorn@tha.de}
}}
\maketitle
\submittednotice
\begin{abstract}
Accurate trajectory prediction is crucial for safe and reliable autonomous driving systems, requiring models that capture long-term temporal dependencies while accounting for social interactions among neighboring vehicles in highway driving scenarios. While Long Short Term Memory (LSTM) networks have been widely used in the domain of trajectory prediction, they have limitations such as limited memory capacity and scalar cell state. The recently introduced Extended Long Short Term Memory (xLSTM) addresses these limitations of traditional LSTMs by introducing exponential gating and enhanced memory structures, making them better suited for modeling long-term temporal dependencies. Despite their potential, xLSTM-based models remain underexplored in the context of vehicle trajectory prediction. This paper introduces a novel xLSTM-based highway trajectory prediction framework, X-TRAJ, as the first application of xLSTM, and its physics-aware variant, X-TRACK (eXtended LSTM for TRAjectory prediction Constraint by Kinematics), which explicitly integrates vehicle motion kinematics into the model learning process. By introducing physical constraints, the proposed model generates realistic and feasible highway trajectories. A comprehensive evaluation on the publicly available highway datasets, highD and NGSIM, demonstrates that X-TRACK outperforms state-of-the-art baselines on highD and is among the state-of-the-art models on the NGSIM dataset.
\end{abstract}
\section{Introduction}
In autonomous driving, anticipating the future trajectories of neighboring vehicles is crucial for safe motion planning \cite{intro_1}. Vehicle motion is influenced not only by individual dynamics but also by social interactions and environmental context. Autonomous vehicles must precisely anticipate the behavior of surrounding vehicles as the environment is highly dynamic, particularly during maneuvers such as lane changes, merging, or braking. Therefore, for accurate trajectory prediction, it is crucial to capture social interactions and, among other factors, the implicit influence of the driving environment reflected in vehicle motion. Additionally, the predicted trajectories must be both socially compliant and physically feasible to ensure the safety of traffic participants.

Due to the ability of Recurrent Neural Networks (RNNs) to model temporal sequences, they are widely used for vehicle trajectory prediction. Particularly, Long Short Term Memory (LSTM) \cite{LSTM} is commonly used in encoder-decoder architectures to capture temporal and social interactions \cite{cs_lstm, mha_lstm, two_channel}. However, purely data-driven RNN-based models often fail to ensure physical feasibility or compliance with physical laws of vehicular motion. Therefore, a kinematic bicycle model was introduced where predictions of a deep learning model are refined through a kinematic layer \cite{dkm}. By incorporating the physical characteristics of the target vehicle, such as non-holonomic dynamics, these hybrid approaches improve trajectory smoothness and feasibility \cite{dkm}. 

Further, conventional LSTMs exhibit several inherent limitations, including restricted memory capacity, scalar-valued cell states, and limited parallelization. To address these limitations, Beck et al. \cite{xLSTM} introduced Extended Long Short Term Memory (xLSTM) with improved memory dynamics, representational capacity, and computational efficiency. 

Recent studies \cite{xLSTM_stock, xLSTM_time, xLSTM_mixer, TiRex} demonstrate the effectiveness of xLSTM in time series forecasting across diverse domains. However, its potential in the field of vehicle trajectory prediction remains underexplored. Richer memory representations and enhanced temporal abstraction make xLSTM well-suited for capturing complex vehicle interactions over extended time frames. Our work leverages the strengths of xLSTM as an encoder to present a highway trajectory prediction architecture, X-TRAJ.

To ensure realistic and physically consistent trajectory predictions, the proposed framework integrates a physics-based kinematic model into the xLSTM-based architecture. While learning-based models achieve high accuracy by modeling motion patterns solely from data, they may produce physically implausible trajectories, such as abrupt turns or unrealistic accelerations \cite{dkm}. In contrast, purely physics-based models strictly adhere to kinematic laws but often lack the flexibility to represent complex traffic behavior \cite{physics}. Explicitly incorporating vehicle dynamics into the data-driven learning process, the proposed hybrid framework, termed X-TRACK (eXtended LSTM for TRAjectory prediction Constraint by Kinematics), leverages the strengths of both paradigms, generating trajectories that are accurate and physically consistent with real-world vehicular motion.

The main contributions are as follows:
\begin{itemize}
    \item A novel trajectory prediction framework that leverages xLSTM's ability to model long-range temporal dependencies for predicting trajectories on highways.
    \item Integration of a physics-based kinematic layer into an xLSTM-based prediction framework to predict trajectories in accordance with the vehicle dynamics.
    \item Evaluation of xLSTM variants for encoding and decoding purposes with respect to the trajectory prediction task.
    \item Performance evaluation on the publicly available highway datasets, highD and NGSIM.
\end{itemize}

\section{Related Work}
Vehicle trajectory prediction aims to forecast the future motion of a target vehicle. In this work, we focus on highway driving scenarios, where motion is primarily influenced by interactions such as car-following, lane keeping, and lane changes. Two major challenges arise in this task: difficulty in modeling social interactions among vehicles and managing the inherent multimodality of future trajectories, where identical histories may lead to multiple feasible trajectories. Prior works range from rule-based models to deep learning approaches. Conventional methods involve maneuver-based, physics-based, and interaction-aware motion models \cite{survey_1}. This section reviews deep learning-based models and physics-based approaches, with a focus on highway trajectory prediction. A comprehensive survey can be found in Lef\`evre et al. \cite{survey_1}.

\subsection{Deep Learning-Based Trajectory Prediction}
Due to the sequential nature of trajectory data, RNNs such as LSTMs \cite{LSTM} and Gated Recurrent Units (GRUs) are widely adopted to model temporal dependencies in vehicular motion. Deo and Trivedi proposed CS-LSTM \cite{cs_lstm}, which uses convolutional social pooling to model vehicle interactions. Messaoud et al. introduced a Multi-Head Attention (MHA) mechanism \cite{mha_lstm} to model distant traffic participants. 

Recent developments leverage Graph Neural Networks (GNNs) to create a scene graph and represent the neighboring participants as nodes. Mo et al. \cite{two_channel} employed Graph Attention Networks (GATs) \cite{gat} to model inter-vehicle interactions, while Repulsion and Attraction Graph Attention (RA-GAT) \cite{ra_gat} captures repulsive and attractive forces within a traffic scenario. Hierarchical Spatio-Temporal Attention (HSTA) \cite{hsta} combines GATs, MHAs, and LSTMs to model spatio-temporal interactions. Chen et al. \cite{inatran} proposed a non-autoregressive transformer-based model for trajectory prediction. Despite the success of LSTMs, they are limited in capturing long-term dependencies. To address this, xLSTM architectures extend traditional LSTMs through exponential gating and augmented memory structures, enabling models to capture long-range temporal dependencies \cite{xLSTM}.

\subsection{Physics-Based Trajectory Prediction}
Incorporating physical constraints leads to the prediction of physically feasible trajectories, which are often overlooked by data-driven deep learning approaches. The kinematic bicycle model is widely used to describe vehicle motion under physical constraints. Cui et al. \cite{dkm} integrated a kinematic layer into a deep learning framework to explicitly encode vehicle dynamics, generating safer and more feasible trajectory predictions. A Similar hybrid approach was adopted by Neumeier et al. \cite{cVMD} to enforce physical consistency by using a kinematic bicycle model in addition to their network. However, when combined with conventional LSTM architectures, such models might still suffer from limited memory capacity due to their scalar cell states. To overcome this, a kinematic layer is combined with an xLSTM architecture that provides enhanced representational capacity and rich memory structures.

\section{xLSTM-based Vehicle Trajectory Prediction}
This section formulates the vehicle trajectory prediction problem and presents the proposed xLSTM-based framework with its components, followed by the integration of a kinematic layer to ensure physically plausible trajectory predictions.

\subsection{Problem Definition}
\label{subsection:problem}
The aim is to predict the future trajectory of the target vehicle on a highway using solely historical trajectories of neighboring vehicles from time $t = 1$ to $t = t_\mathrm{obs}$. The input to the model is the history of the target and its $N$ surrounding vehicles in the same lane as the target vehicle and the lanes adjacent to the target vehicle lane. The input trajectory of a vehicle $i$ is defined as $\mathbf{X}_i = [\,\mathbf{x}_i^1, \mathbf{x}_i^2, \dots, \mathbf{x}_i^{t_\mathrm{obs}}\,]$ where $\mathbf{x}_i^t = [\,x_i^t, y_i^t, v_i^t, a_i^t\,]^\top$. Here, $x_i^t$ and $y_i^t$ denote the position coordinates of the vehicle $i$ at time $t$, and $v_i^t$, $a_i^t$ are the velocity and acceleration, respectively.

The position coordinates of all the vehicles are represented in a stationary frame of reference with the origin fixed at the target vehicle's position at time $t=1$. The $x-axis$ points towards the direction of motion on the highway, and the $y-axis$ points downward, perpendicular to the $x-axis$.

X-TRAJ outputs the predicted positions of the target vehicle for the next $t_f$ time steps. 
\begin{equation}
     \mathbf{Y}_\mathrm{pred} = [\,\mathbf{y}_\mathrm{pred}^{t_\mathrm{obs}+1}, \mathbf{y}_\mathrm{pred}^{t_\mathrm{obs}+2}, \dots, \mathbf{y}_\mathrm{pred}^{t_\mathrm{obs}+t_f}\,],
\end{equation}
where $\mathbf{y}_\mathrm{pred}^t = [\ x_\mathrm{pred}^t, y_\mathrm{pred}^t\,]^\top$ is the predicted position coordinates of the target vehicle at time $t$.

For X-TRACK, the input of the model is defined in a similar way as $\mathbf{X}_i = [\,\mathbf{x}_i^1, \mathbf{x}_i^2, \dots, \mathbf{x}_i^{t_\mathrm{obs}}\,]$, where $\mathbf{x}_i^t = [\, a_{x, i}^t, \dot{\psi}_i^t\,]^\top$, where longitudinal acceleration ($a_x$) and yaw rate ($\dot{\psi}$) represents the vehicle's kinematic state. Similarly, the model predicts $\mathbf{y}_\mathrm{pred}^t = [\, a_{x,\mathrm{pred}}^t,\, \dot{\psi}_\mathrm{pred}^t\,]^\top$ i.e., the longitudinal acceleration and yaw rate for the next time steps.

\subsection{Overview of xLSTM Architecture}
Beck et al. introduced xLSTM \cite{xLSTM} with two novel building blocks: sLSTM and mLSTM, to overcome the limitations of original LSTMs \cite{LSTM}. sLSTM has a scalar memory, scalar updates, and memory mixing, whereas mLSTM has matrix memory and a covariance update rule. In our trajectory prediction framework, an sLSTM block is employed as the encoder. Exponential gating combined with normalization and stabilization is used to provide sLSTM the ability to revise storage decisions. In addition to this, sLSTM can have multiple heads that enable memory mixing via recurrent connections $R_x$ where $x \in \{i, f, o\}$ from the hidden state vector to memory cell input $z$ and the gates $i, f, o$. This new way of memory mixing is possible within each head but not across heads. The forward pass of sLSTM is given by \cite{xLSTM}:
\begin{align}
\mathbf{c}_t &= \mathbf{f}_t \odot \mathbf{c}_{t-1} + \mathbf{i}_t \odot \mathbf{z}_t, \displaybreak[1] \notag\\[6pt]
\mathbf{n}_t &= \mathbf{f}_t \odot \mathbf{n}_{t-1} + \mathbf{i}_t, \quad \displaybreak[1] \notag\\[6pt]
\mathbf{h}_t &= \mathbf{o}_t \odot \tilde{\mathbf{h}}_t, \quad \tilde{\mathbf{h}}_t = {\mathbf{c}_t}\odot {\mathbf{n}_t}^{-1}, \displaybreak[1] \notag\\[6pt]
\mathbf{z}_t &= \varphi(\tilde{\mathbf{z}}_t), \quad \tilde{\mathbf{z}}_t = \mathbf{W}_z^\top \mathbf{x}_t + \mathbf{R}_z \mathbf{h}_{t-1} + \mathbf{b}_z,  \displaybreak[1] \notag\\[6pt]
\mathbf{i}_t &= \exp(\tilde{\mathbf{i}}_t), \quad \tilde{\mathbf{i}}_t = \mathbf{W}_i^\top \mathbf{x}_t + \mathbf{R}_i \mathbf{h}_{t-1} + \mathbf{b}_i, \displaybreak[1] \notag\\[6pt]
\mathbf{f}_t &= \sigma(\tilde{\mathbf{f}}_t) \;\; \text{or} \;\; \exp(\tilde{\mathbf{f}}_t), 
\quad \tilde{\mathbf{f}}_t = \mathbf{W}_f^\top \mathbf{x}_t + \mathbf{R}_f \mathbf{h}_{t-1} + \mathbf{b}_f, \displaybreak[1]  \notag\\[6pt]
\mathbf{o}_t &= \sigma(\tilde{\mathbf{o}}_t), \quad \tilde{\mathbf{o}}_t = \mathbf{W}_o^\top \mathbf{x}_t + \mathbf{R}_o \mathbf{h}_{t-1} + \mathbf{b}_o, 
\label{eq:xlstm}
\end{align}
where $\mathbf{c}_t, \mathbf{n}_t, \mathbf{h}_t \in \mathbb{R}^d$ represent the cell state, normalizer state, and hidden state, respectively, and $\odot$ is element-wise multiplication. $\varphi$ is the cell input activation function, typically $tanh$, $\sigma$ is sigmoid and $\exp$ is exponential activation function. $\mathbf{W}_z, \mathbf{W}_i, \mathbf{W}_f, \mathbf{W}_o$ are the input weight matrices between inputs $\mathbf{x}_t$ and cell input, input gate, forget gate, and output gate, respectively. $\mathbf{R}_z, \mathbf{R}_i, \mathbf{R}_f, \mathbf{R}_o$ correspond to the recurrent weights between hidden state $\mathbf{h}_{t-1}$ and cell input, input gate, forget gate, and output gate, respectively. The corresponding bias terms are $\mathbf{b}_z, \mathbf{b}_i, \mathbf{b}_f, \mathbf{b}_o$. A detailed discussion of (\ref{eq:xlstm}), as well as comparison to LSTMs, can be found in \cite{xLSTM}.

\subsection{Overall Framework}

\begin{figure*}[htbp]
    \centering
    \includegraphics[width=1.02\textwidth]{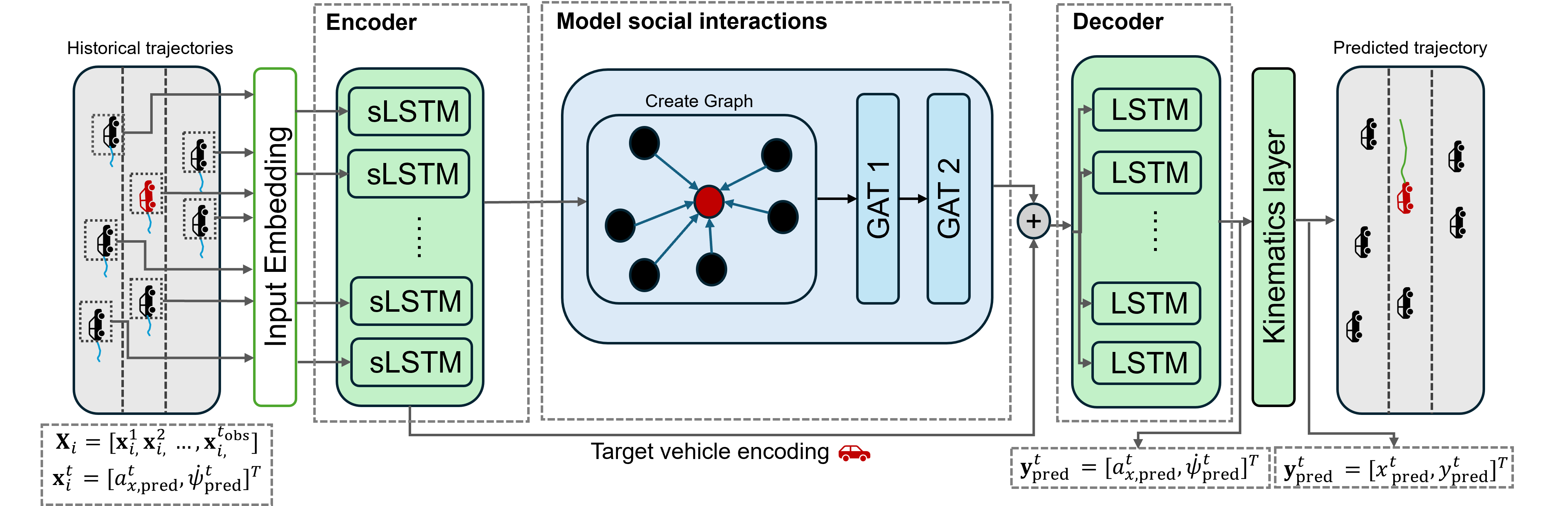}
    \caption{\textbf{Proposed X-TRACK architecture}: The sLSTM block generates an encoded vector for all the vehicles in the traffic scenario. The GAT layers model the vehicle interactions between the target vehicle (shown in red) and the neighboring vehicles based on attention scores. The concatenation of the output from the GAT module and target vehicle encoding is passed through the decoder to predict future motion parameters. The Kinematic layer then transforms these motion parameters into position coordinates to provide the future trajectory of the target vehicle.}
    \label{fig:xLSTM_architecture}
\end{figure*}

This section defines the proposed trajectory prediction framework, illustrated in Fig. \ref{fig:xLSTM_architecture}. The model features an encoder-decoder structure, where an sLSTM encoder processes the sequential position of each vehicle to represent the temporal evolution of trajectories. sLSTM extends vanilla LSTM with exponential gating, scalar memory updates, and recurrent memory mixing, enabling sequential state tracking as evident in \cite{xLSTM}. In contrast, mLSTM employs parallelizable matrix memory without hidden-to-hidden recurrence, enabling scalability while trading sequential mixing. Additionally, sLSTM was employed by Kraus et al. \cite{xLSTM_mixer} in xLSTM-Mixer for multivariate time series to capture complex temporal relationships, highlighting the unsuitability of mLSTM due to their independent treatment of the sequence elements \cite{xLSTM_mixer}, making sLSTM better suited for trajectory prediction in our setting.

To model social interactions, the encoded vectors are forwarded to a GAT to compute the attention scores among vehicles and create a context vector. The future trajectory of the target vehicle, conditioned on the target encoding and social context, is then predicted using an LSTM decoder. The LSTM decoder ensures stable and efficient autoregressive predictions, as more complex variants such as sLSTM did not produce additional benefits for short-horizon prediction. This design choice of using an sLSTM as the encoder and an LSTM as the decoder is motivated by their theoretical properties and subsequently validated through empirical evaluation in Section \ref{sec:ablation}. Our architecture leverages the strengths of social modeling, first introduced by Mo et al. \cite{two_channel}, and differs in two significant aspects: an xLSTM encoder for improved modeling of long-term dependencies, and the integration of a physics-based kinematics model to generate physically plausible trajectories.

\subsubsection{Temporal \textbf{Encoder}} This layer encodes the trajectories of the target vehicle and its $N = 8$ neighboring vehicles: preceding, following, left/right preceding, left/right alongside, left/right following. For every neighboring vehicle $i$, the state vector at time $t$, denoted as $\mathbf{x}_i^t$, is passed through a fully connected layer to create an embedding $\mathbf{e}_i^t$:
\begin{equation}
    \mathbf{e}_i^t = \Psi(\mathbf{x}_i^t; \mathbf{W}_{\mathrm{emb}}),
\end{equation}
where $\Psi(.)$ is implemented as a fully connected layer with a LeakyReLU activation, and $\mathbf{W}_\mathrm{emb}$ represents the learnable weights. xLSTM encoder captures the temporal dynamics by processing these embedding vectors sequentially. 
\begin{equation}
    \mathbf{h}_i^t = \operatorname{xLSTM}(\mathbf{h}_i^{t-1}, \mathbf{e}_i^t; \mathbf{W}_\mathrm{enc}),
\end{equation}
where $\mathbf{h}_i^t$ denotes the hidden state of the vehicle $i$ at time $t$ and $\mathbf{W}_\mathrm{enc}$ are the encoder weights. In the case of the encoder, an sLSTM block is employed.
\subsubsection{Vehicle Interaction Modeling} 
The interaction among vehicles is modeled as a graph where each vehicle $i$ in the scenario is represented as a node in a graph, with the central node being the target vehicle. The central node is indexed at $0$, and all the neighboring vehicles are indexed at ${1, \dots, N}$.
\begin{equation}
    \mathbf{E} = \{\,e^{0,j}\,\}_{j=1}^{N},
\end{equation}
where $e^{0, j}$ is a directed edge from node $j$ to node $0$ (the target vehicle node) which implies that node $j$ is a neighbor of target node and node $0$'s behavior is impacted by node $j$'s behavior.

At each time $t$, node features are the xLSTM encoder states $\mathbf{h}_i^t \in \mathbb{R}^d$. A two-layer multi-head GAT model is used to predict the pairwise vehicular interactions within the traffic scenario. The hidden state $\mathbf{h}_i^t$ is passed through two successive GAT \cite{gat} layers $\mathcal{G}_1(.)$ and $\mathcal{G}_2(.)$ that compute attention scores over neighbors and aggregate their messages; outputs $\tilde{\mathbf{z}}_T^t$ of the target vehicle $T$ are projected by a linear layer to produce an interaction feature $\mathbf{g}_T^t$. 
\begin{align}
\mathbf{z}_i^t &= \mathcal{G}_1(\mathbf{h}_i^t; \mathbf{W}_{\mathrm{gat1}}), \notag\\
\tilde{\mathbf{z}}_i^t &= \mathcal{G}_2(\mathbf{z}_i^t; \mathbf{W}_{\mathrm{gat2}}), \notag\\
\mathbf{g}_i^t &= \Phi(\tilde{\mathbf{z}}_i^t; \mathbf{W}_{\mathrm{fc}}),
\label{eq:gat_module}
\end{align}
where $\mathbf{W}_\mathrm{gat1}$ and $\mathbf{W}_\mathrm{gat2}$ are the learnable weights of GAT layers. $\Phi(.)$ is a fully connected layer with LeakyReLU activation and $\mathbf{W}_{\mathrm{fc}}$ are the learned projection weights.
\subsubsection{Future \textbf{Decoder}} \label{subsubsection:xLSTM} LSTM decoder predicts the future trajectory of the target vehicle T using the concatenation of the target vehicle's encoding vector $\mathbf{h}_T^t$ and interaction vector $\mathbf{g}_T^t$ generated by the GAT module. The LSTM layer is followed by a fully connected layer that outputs the future position coordinates of the target vehicle for time steps $t = t_\mathrm{obs}+1, t_\mathrm{obs}+2, \dots, t_\mathrm{obs}+t_f$:
\begin{equation}
    \mathbf{y}_\mathrm{pred}^t = \Lambda\!\Big(\operatorname{LSTM}\big([\mathbf{h}_T^t ; \mathbf{g}_T^t],\, \mathbf{h}_\mathrm{dec}^{t-1};\, \mathbf{W}_\mathrm{dec}\big)\Big),
\end{equation}
where $\mathbf{y}_t$ is the predicted position of the target vehicle at time $t$. The mapping function $\Lambda(.)$ consists of a fully connected layer followed by LeakyReLU non-linearity. The decoder is parameterized by the learnable weights $\mathbf{W}_\mathrm{dec}$, and the hidden state vector from the previous time step is $\mathbf{h}_\mathrm{dec}^{t-1}$.
\subsubsection{Physics-based Kinematic Layer} \label{subsubsection:xLSTM-Kinematic}
Instead of directly predicting the position coordinates, the future decoder predicts the motion parameters, namely the yaw rate $\dot{\psi}^t$ and longitudinal acceleration $a_x^t$, as described in \ref{subsection:problem}, which are subsequently transformed into trajectory coordinates using the physics-based kinematic layer. The encoder-decoder structure remains the same; however, the input to the physics-aware variant, X-TRACK, is $\mathbf{X}_i = [\,\mathbf{x}_i^1, \mathbf{x}_i^2, \dots, \mathbf{x}_i^{t_\mathrm{obs}}\,]$, where $\mathbf{x}_i^t = [\, a_{x, i}^t, \dot{\psi}_i^t\,]^\top$. 

To compute ground truth, numerical derivatives $\bm{a}_x = \frac{d^2 \bm{x}_{\text{pred}}}{dt^2}$ and $\dot{\bm{\psi}} = \arctan\!\left(\frac{d\bm{y}_{\text{pred}}}{dt}, \frac{d\bm{x}_{\text{pred}}}{dt}\right)$ are calculated. From the predicted $\mathbf{Y}_\mathrm{pred} = [\,\mathbf{y}_\mathrm{pred}^{t_\mathrm{obs}+1}, \mathbf{y}_\mathrm{pred}^{t_\mathrm{obs}+2}, \dots, \mathbf{y}_\mathrm{pred}^{t_\mathrm{obs}+t_f}\,],$ where $\mathbf{y}_\mathrm{pred}^t = [\,a_{x,\mathrm{pred}}^t,\, \dot{\psi}_\mathrm{pred}^t\,]^\top$, the position $[\ x^t, y^t\,]^\top$, velocity $v^t$ and heading angle $\psi^t$ of a vehicle at time step $t+\Delta t$ are computed as shown below \cite{botsch_utschick, cVMD}:
\begin{align}
x^{t+\Delta t} &= x^t + v^t c(\psi^t)\,\Delta t 
+ \Big(a_x^t c(\psi^t) - \dot{\psi}^t v^t s(\psi^t)\Big)\frac{{\Delta t}^2}{2}, \nonumber \\
y^{t+\Delta t} &= y^t + v^t s(\psi^t)\,\Delta t 
+ \Big(a_x^t s(\psi^t) + \dot{\psi}^t v^t c(\psi^t)\Big)\frac{{\Delta t}^2}{2}, \nonumber \\
v^{t+\Delta t} &= v^t + a_x^t\,\Delta t, \nonumber \\
\psi^{t+\Delta t} &= \psi^t + \dot{\psi}^t\,\Delta t, \label{eq:kinematic}
\end{align}
where $c(\psi^t) = \cos(\psi^t),  \quad s(\psi^t) = \sin(\psi^t)$, $\psi^t$ being the heading angle at time $t$ and $\Delta t$ is the time step size.

The predicted motion parameters are bounded by the physical limits $a_{x,\max} = \pm 9 \; \text{m/s}^2 \;$ \cite{Yusof2016} and $\dot{\psi}_{\max} = \pm 71.26 \; \text{deg/s}$ \cite{Kontos2023} to ensure that the predicted trajectory is feasible to be executed by the target vehicle. X-TRACK does not rely on position coordinates to understand the vehicle dynamics; instead, it uses non-holonomic constraints of the vehicle to predict reliable future motion. 
\section{Dataset, Experiments and Evaluation}
\subsection{Datasets}
\label{subsection:preprocessing}
To ensure a fair evaluation of the proposed framework on highway data, the publicly available Highway Drone (highD) and Next Generation SIMulation (NGSIM) datasets are used. 

\textbf{highD:} The dataset has vehicle trajectories with a sample rate of $f = 25$ Hz captured using a drone at six different highway locations in Germany. There is a large imbalance in scenarios where keep lane scenarios are more frequent than lane change scenarios \cite{GFTNNV2}. To address this, traffic scenarios are extracted during the preprocessing step to ensure uniform distribution. Later, the extracted scenarios are split into subsets with $70:10:20$ ratio: train, validation, and test with $9604$, $1371$, $2747$ samples, respectively.

\textbf{NGSIM:} The dataset was recorded on US highways US-$101$, I-$80$ with a sample rate of $f = 10$ Hz. Similar to highD, NGSIM also has a huge imbalance of scenarios \cite{ra_gat}. Therefore, the data is preprocessed in a similar way to extract an equal proportion of each scenario type. This, in turn, heavily reduces the total number of scenarios in the NGSIM dataset; however, training on a balanced dataset is crucial to avoid biasing the model towards lane-keeping scenarios, which constitute the majority class in both datasets \cite{GFT}. The proportionate dataset is then split into train, validation, and test sets with $1634$, $232$, and $471$ scenarios, respectively. 

Each scenario spans $8$s, with $3$s of observed history and $5$s of future trajectory. In scenarios with insufficient surrounding context, such as when only a few surrounding vehicles are present, ghost vehicles are introduced during X-TRAJ training, following the successful strategy adopted in prior work \cite{GFT} to handle missing interaction participants. These vehicles are provided with the same motion features as the target vehicle, to avoid adding artificially simulated dynamic \cite{GFT}.

\subsection{Training and Implementation Details}
The model is implemented in PyTorch \cite{pytorch} with GAT layers using PyTorch Geometric \cite{pyg}. The encoder consists of a single-layer sLSTM block \cite{xLSTM} with a $64$-dimensional hidden state and the LSTM decoder with a $128$-dimensional hidden state. The input features are transformed into a $32$-dimensional space before encoding. The GAT layer \cite{gat} employs four concatenated attention heads, and LeakyReLU with a negative slope of $0.1$ is used. The model is trained using a batch size of $32$ with the Adam optimizer \cite{adam}. Training is performed using a combined trajectory and kinematic loss, where $\mathcal{L}_{\text{traj}}$ is a weighted mean squared error over the prediction horizon and $\mathcal{L}_{\text{dyn}}$ supervises the predicted longitudinal acceleration and yaw rate.
\begin{align}
&\mathcal{L} = \mathcal{L}_{\text{traj}} + \alpha_{\text{dyn}} \mathcal{L}_{\text{dyn}}, \nonumber \\
&\mathcal{L}_{\text{traj}} = \beta_x \mathrm{MSE}(\bm{x}_{\text{pred}}, \bm{x}_{\text{gt}}) + \beta_y \mathrm{MSE}(\bm{y}_{\text{pred}}, \bm{y}_{\text{gt}}), \nonumber \\
&\mathcal{L}_{\text{dyn}} = \mathrm{MSE}(\bm{a}_{x, \text{pred}}, \bm{a}_{x, \text{gt}}) + \mathrm{MSE}(\bm{\dot{\psi}}_{\text{pred}}, \bm{\dot{\psi}}_{\text{gt}}),
\end{align}
where $\alpha_{\text{dyn}}$ balances the influence of kinematic supervision: $\alpha_{\text{dyn}} = 0$ for X-TRAJ training (kinematic module disabled) and $\alpha_{\text{dyn}} = 0.1$ for X-TRACK training. The value of $\alpha_{\text{dyn}} = 0.1$ was selected through empirical evaluation as it achieved optimal performance by balancing kinematic constraints with trajectory accuracy without allowing the dynamic loss to dominate. Empirically, $\beta_x = 20$ and $\beta_y = 0.5$ yielded the best results and are consistent with prior findings in \cite{two_channel}. For code and further implementation details, see the link to \href{https://github.com/archugh02/X-TRACK}{the GitHub repository}.
\subsection{Evaluation Metrics}
The following commonly used metrics are used to report and evaluate our model as compared to baselines: \\
Average Displacement Error (ADE): The mean Euclidean distance between the predicted and ground truth trajectories averaged across all time steps and all trajectories. 
{\small
\begin{align}
\mathrm{ADE} =
\frac{1}{N\, t_f}
\sum_{n=1}^{N}
\sum_{t=t_{\mathrm{obs}}+1}^{t_{\mathrm{obs}}+t_f}
\sqrt{
\big(x_{\mathrm{gt},n}^{t} - x_{\mathrm{pred},n}^{t}\big)^{2} +
\big(y_{\mathrm{gt},n}^{t} - y_{\mathrm{pred},n}^{t}\big)^{2}
} \label{eq:ade}
\end{align} 
}\\
Final Displacement Error (FDE): The Euclidean distance between the predicted and ground truth final positions for each trajectory, averaged over all trajectories. 
{\small
\begin{align}
\mathrm{FDE} = 
\frac{1}{N}
\sum_{n=1}^{N}
\sqrt{
\left(x_{\mathrm{gt},n}^{t_\mathrm{obs} + t_f} - x_{\mathrm{pred},n}^{t_\mathrm{obs} + t_f}\right)^{2} +
\left(y_{\mathrm{gt},n}^{t_\mathrm{obs} + t_f} - y_{\mathrm{pred},n}^{t_\mathrm{obs} + t_f}\right)^{2}
} \label{eq:fde}
\end{align}
}\\
Root Mean Square Error (RMSE) at time t: The square root of the average of the squared differences between the predicted and corresponding ground truth positions for all $N$ trajectories.
{\footnotesize
\begin{align}
\mathrm{RMSE}(t) = 
\sqrt{
\frac{1}{N} \sum_{n=1}^{N} \Big[
    \left(x_{\mathrm{gt},n}^{t} - x_{\mathrm{pred},n}^{t}\right)^{2}
    + 
    \left(y_{\mathrm{gt},n}^{t} - y_{\mathrm{pred},n}^{t}\right)^{2}
\Big]
}
\label{eq:rmse}
\end{align}
}
\subsection{Comparison with Baselines}
For evaluation, the model's performance is benchmarked against state-of-the-art LSTM-based, Transformer-based, and Diffusion-based kinematic trajectory prediction methods. Within the scope of this work, only models that rely solely on highway trajectory data, without map or road-structure priors, are considered. Below are the models considered in comparison:
\begin{itemize}
    \item \textbf{Convolutional Social LSTM (CS-LSTM)} \cite{cs_lstm}: To model vehicle interaction, CS-LSTM uses convolution layers with social pooling and an LSTM-based encoder-decoder.
    \item \textbf{Graph and Recurrent Neural Network (Two Channel)} \cite{two_channel}: Utilizes GATs to model social interactions with a GRU encoder and LSTM decoder to predict the future path of the vehicle. 
    \item \textbf{Multi-Head Attention LSTM (MHA-LSTM)} \cite{mha_lstm}: an LSTM-based encoder-decoder model that leverages both global and local attention to capture social interactions between the target and its neighboring vehicles.
    \item \textbf{Repulsion and Attraction Graph Attention (RA-GAT)} \cite{ra_gat}: Model vehicles and spaces as nodes in a graph and employ a dual graph attention mechanism to model repulsion from surrounding vehicles and attraction to the non-vehicle spaces.
    \item \textbf{Hierarchical Spatio-Temporal Attention (HSTA)} \cite{hsta}: MHA is used to encode temporal correlations of interactions along with GAT to capture spatial interactions, followed by a state-gated fusion layer to integrate both spatial and temporal dependencies. 
    \item \textbf{Intention-aware Non-Autoregressive Transformer (iNATran)} \cite{inatran}: Integrates graph attention with the transformer encoder and temporal attention learning (TAL) module to capture social and temporal dependencies. In the decoder, it combines cross-attention learning with intention-aware query generation.
    \item \textbf{Graph Fourier Transformation Neural Network (GFTNNv2)} \cite{GFTNNV2}: Vehicle interactions are transformed using Graph Fourier Transform (GFT) into a spectral scenario representation. This is fed to a neural network, and the prediction is converted to the spatio-temporal domain by applying the inverse GFT.
    \item \textbf{Conditioned Vehicle Motion Diffusion (cVMD)} \cite{cVMD}: The framework uses diffusion models and integrates a kinematic layer \cite{botsch_utschick} (physical and non-holonomic motion constraints) with an uncertainty quantification module to enhance its prediction performance.
    \item \textbf{Graph-Based Spatial-Temporal Attentive Network (GSTAN)} \cite{GSTAN}: To efficiently capture vehicle interactions, this model leverages attentive spatio-temporal modeling. The weighted distance GATs are integrated with dynamic attention assignment and modeled using multi-headed attention-based transformers, which capture social interaction patterns.
    \item \textbf{X-TRAJ and X-TRACK}: The proposed model in this paper, where xLSTM is used to encode relationships with LSTM as a decoder and integrate it with the kinematic bicycle model to generate feasible trajectories. The two proposed models: one without the kinematic layer (X-TRAJ, see \ref{subsubsection:xLSTM}) and another with the kinematic layer (X-TRACK, see \ref{subsubsection:xLSTM-Kinematic}).
\end{itemize}
\section{Results and Discussion}
\begin{table}[!htbp]
\caption{ADE (in meters) and FDE (in meters) evaluated at a $5$s prediction horizon for the models on highD and NGSIM datasets}
\centering
\renewcommand{\arraystretch}{1.15}
\begin{tabular}{|
p{0.27\columnwidth} |
p{0.11\columnwidth} |
p{0.11\columnwidth} |
p{0.11\columnwidth} |
p{0.11\columnwidth} |}
\hline
\textbf{Architecture} & 
\multicolumn{2}{c}{\textbf{highD}} & 
\multicolumn{2}{|c|}{\textbf{NGSIM}} \\
\cline{2-5} 
 & \textbf{{ADE}} & \textbf{{FDE}} & 
   \textbf{{ADE}} & \textbf{{FDE}} \\
\hline
X-TRACK (Ours)         & \textbf{0.56} & \textbf{1.76} & 2.11 & 5.17 \\
\hline
X-TRAJ (Ours)      & 1.14  & 2.65 & \underline{1.99} & 4.99 \\
\hline
CS-LSTM \cite{cs_lstm}   & 2.14 & 5.09 & 2.69 & 5.81 \\
\hline
Two Channel \cite{two_channel}    & 2.08 & 4.84 & 2.89 & 6.26 \\
\hline
MHA-LSTM \cite{mha_lstm}  & 1.97 & 4.72 & 2.54 & 5.47 \\
\hline
RA-GAT \cite{ra_gat}        & 1.99 & 4.83 & 2.70 & 5.88 \\
\hline
HSTA \cite{hsta}  & 1.99 & 4.60 & 2.54 & 5.80 \\
\hline
iNATran \cite{inatran}  & 1.84 & 3.95 & 2.43 & 5.33 \\
\hline
GFTNNv2 \cite{GFTNNV2}    & \underline{0.92} & \underline{2.20} & 2.26 & \underline{4.83} \\
\hline
cVMD \cite{inatran}  & 1.74 & 4.95 & 3.02 & 7.23 \\
\hline
GSTAN \cite{GSTAN}        & 1.28 & 2.66 & \textbf{1.94} & \textbf{4.50} \\
\hline
\multicolumn{5}{l}{\footnotesize Note: The best results are in \textbf{bold}, and the second-best are \underline{underlined}.}
\end{tabular}
\label{tab: ade}
\end{table}
\begin{table}[!htbp]
\caption{RMSE in meters over a 5-second prediction horizon for the models on highD and NGSIM datasets}
\begin{minipage}{0.48\textwidth}
\begin{center}
\renewcommand{\arraystretch}{1.3}
\setlength{\tabcolsep}{5pt}
\begin{tabular}{|l|l|c|c|c|c|c|}
\hline
\textbf{Dataset} & \textbf{Architecture} & \multicolumn{5}{c|}{\textbf{Prediction Horizon (s)}} \\
\cline{3-7} 
 &  & \textbf{1} & \textbf{2} & \textbf{3} & \textbf{4} & \textbf{5} \\
 \hline
\multirow{10}{*}{highD} 
 & X-TRACK (Ours)     & \textbf{0.10} & \textbf{0.31}  & \textbf{0.71}  & \textbf{1.31}  & \textbf{2.16}  \\
 & X-TRAJ (Ours)                &  0.48 & 1.01  & 1.58  & 2.20  & 3.17  \\
 & CS-LSTM \cite{cs_lstm}   & 0.76  & 1.77  & 3.08  & 4.66  & 6.52  \\
 & Two Channel \cite{two_channel} & 0.81  &  1.70 & 2.94  & 4.41  & 6.14  \\
 & MHA-LSTM \cite{mha_lstm} & 0.71  & 1.62  & 2.85  & 4.31  & 6.06  \\
 & RA-GAT \cite{ra_gat}     &  0.67 & 1.58  & 2.79  & 4.29  & 6.06  \\
 & HSTA \cite{hsta}         & 1.16  & 2.43  & 3.84  & 5.38  & 7.09  \\
 & iNATran \cite{inatran}         & 0.88 & 1.62  & 2.35  & 3.58  & 4.95 \\
 & GFTNNv2 \cite{GFTNNV2}       & 0.47  & \underline{0.61}  & \underline{1.05}  & \underline{1.75}  &  \underline{2.69} \\
 & cVMD \cite{cVMD}         & \underline{0.26}  & 1.03  & 2.27  & 3.93  & 5.99 \\
 & GSTAN \cite{GSTAN}       & 0.51  & 1.02  & 1.60  & 2.23 & 3.09  \\
\hline
\multirow{10}{*}{NGSIM} 
 & X-TRACK (Ours)     & \textbf{0.68}  & 1.78  & 3.25  & 4.80 & 6.67  \\
 & X-TRAJ (Ours)               & \underline{0.73}  & \textbf{1.65}  & \underline{2.84}  &  \underline{4.11} & 6.44  \\
 & CS-LSTM \cite{cs_lstm}   & 1.17  & 2.24  & 3.84  & 5.64  & 7.66  \\
 & Two Channel \cite{two_channel} & 1.23  & 2.50  & 4.17  & 6.04  & 8.15  \\
 & MHA-LSTM \cite{mha_lstm} & 1.17  & 2.14  & 3.60  & 5.22  & 7.11  \\
 & RA-GAT \cite{ra_gat}     & 1.24  & 2.24  & 3.80  & 5.54  & 7.65  \\
 & HSTA \cite{hsta}         & 0.97  & 2.11  & 3.66  & 5.35  & 7.40  \\
 & iNATran \cite{inatran}         &  1.20 & 2.07  & 3.51  & 5.05  & 7.03 \\
 & GFTNNv2 \cite{GFTNNV2}       & 0.99  & 1.98  & 3.26  & 4.67  &  \underline{6.30} \\
 & cVMD \cite{cVMD}         & 0.92  &  2.38 & 4.26  & 6.43  & 8.88 \\
 & GSTAN \cite{GSTAN}       & 0.89  & \underline{1.66}  & \textbf{2.74}  & \textbf{4.00}  & \textbf{5.90}  \\
\hline
\multicolumn{7}{l}{\footnotesize Note: The best results are in \textbf{bold}, and the second-best are \underline{underlined}.}
\end{tabular}
\label{tab: pred}
\end{center}
\end{minipage}
\end{table}
Table \ref{tab: ade} reports the ADE and FDE values for compared models on highD and NGSIM datasets, and Table \ref{tab: pred} summarizes the RMSE errors over the 5-second prediction horizon. All the models were implemented (in case the code was not available), trained, and evaluated on the same balanced dataset resulting from the data preprocessing step as described in \ref{subsection:preprocessing}. The dataset is adjusted to accommodate the input required by the evaluated models. Both the models proposed in this paper, X-TRAJ and X-TRACK, achieve the best performance across all evaluation metrics in highD and performance comparable to state-of-the-art baselines in the case of NGSIM.

\begin{figure}[t]
    \centering
    \begin{minipage}[t]{0.95\linewidth}
        \centering
        \includegraphics[width=\linewidth]{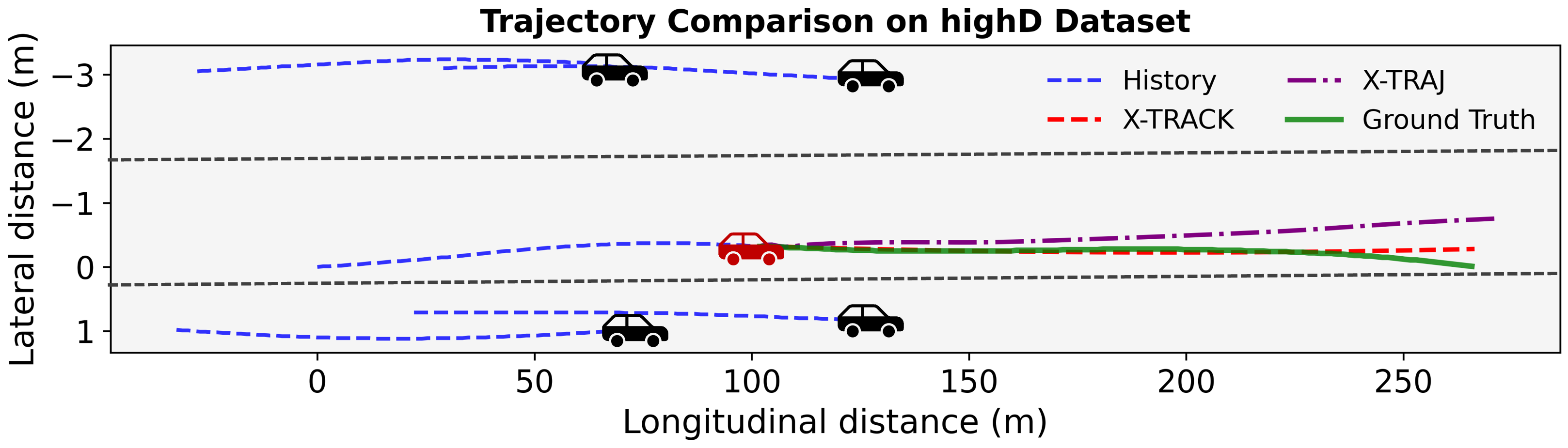}
        \\[3pt] {\scriptsize (a) Keep Lane scenario}
    \end{minipage}
    \\[8pt]
    \begin{minipage}[t]{0.95\linewidth}
        \centering
        \includegraphics[width=\linewidth]{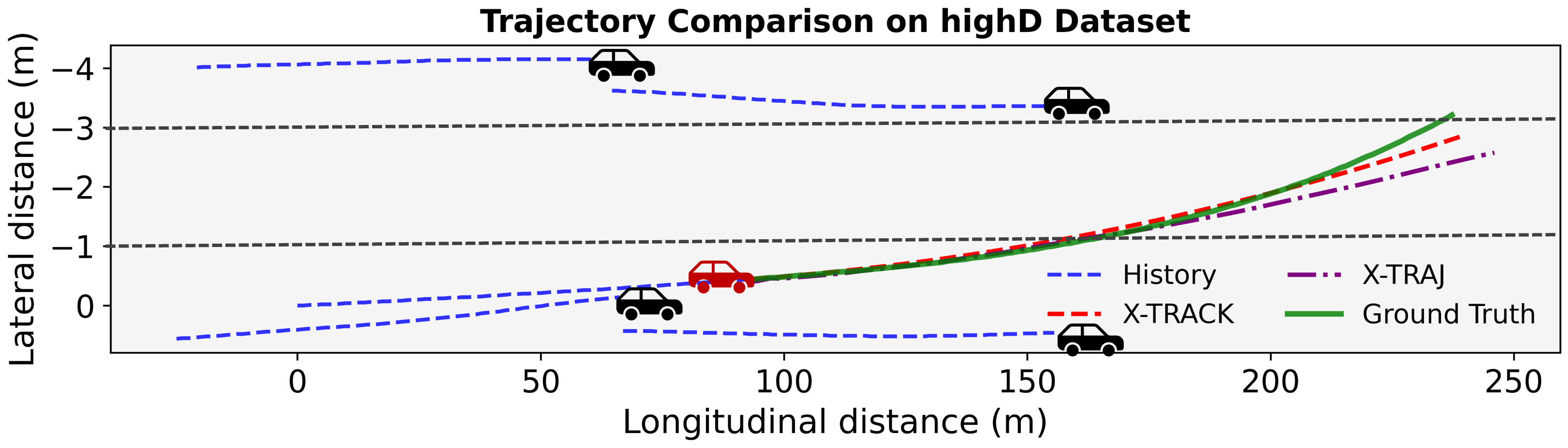}
        \\[3pt] {\scriptsize(b) Lane Change scenario}
    \end{minipage}
    \caption{Comparison of predicted trajectories on the highD dataset using X-TRAJ and X-TRACK in case of (a) Keep lane (b) Lane change.}
    \label{fig:traj_compare}
\end{figure}

On highD, the integration of the kinematic layer to our xLSTM-based vehicle trajectory prediction framework yields substantial improvements: $51\%$ in ADE and $34\%$ in FDE. For RMSE, the X-TRACK model improves by $79\%$ at the $1$-second prediction and by $32\%$ at the $5$-second prediction. This performance enhancement indicates that the non-holonomic constraints introduced by the kinematic layer enforce physically consistent motion and prevent implausible trajectory drift. Fig. \ref{fig:traj_compare} presents two representative prediction examples from the highD dataset, comparing X-TRAJ and X-TRACK, and shows that X-TRACK exhibits closer alignment with the ground-truth trajectory. 

Quantitatively, X-TRACK outperforms the compared baselines on the highD dataset across the prediction horizons. Compared to the best performing baseline without xLSTM (GFTNNv2 \cite{GFTNNV2}), X-TRACK achieves $78.7\%$ and $19.7\%$ improvement at $1$-second and $5$-second prediction, respectively. In contrast, X-TRAJ performs $2.13\%$ and $17.84\%$ worse than GFTNNv2 at the $1$ and $5$-second prediction, respectively, on the highD dataset, highlighting the importance of incorporating kinematic constraints.

For NGSIM, X-TRAJ achieves better performance, with $5.69\%$ and $3.48\%$ improvements in ADE and FDE, respectively, compared to X-TRACK. The second-best performance is achieved by X-TRAJ, preceded by GSTAN \cite{GSTAN}, with $2.51\%$ and $9.82\%$ improvements in ADE and FDE, respectively, relative to X-TRAJ. In terms of RMSE, the lowest RMSE at $1$-second is achieved by X-TRACK, and at $5$-second is achieved by GSTAN. X-TRACK shows $23.6\%$ better performance than GSTAN \cite{GSTAN} at $1$-second RMSE and $13.05\%$ worse performance at $5$-second RMSE.

Compared to highD, all evaluated models exhibit higher ADE, FDE, and RMSE on NGSIM, with performance gains being more pronounced on highD. This behavior highlights a well-established characteristic of physics-based trajectory prediction: non-holonomic constraints are most effective when the underlying data is physically consistent and smoothly annotated. 

\begin{figure}
    \centering
    \includegraphics[width=\linewidth]{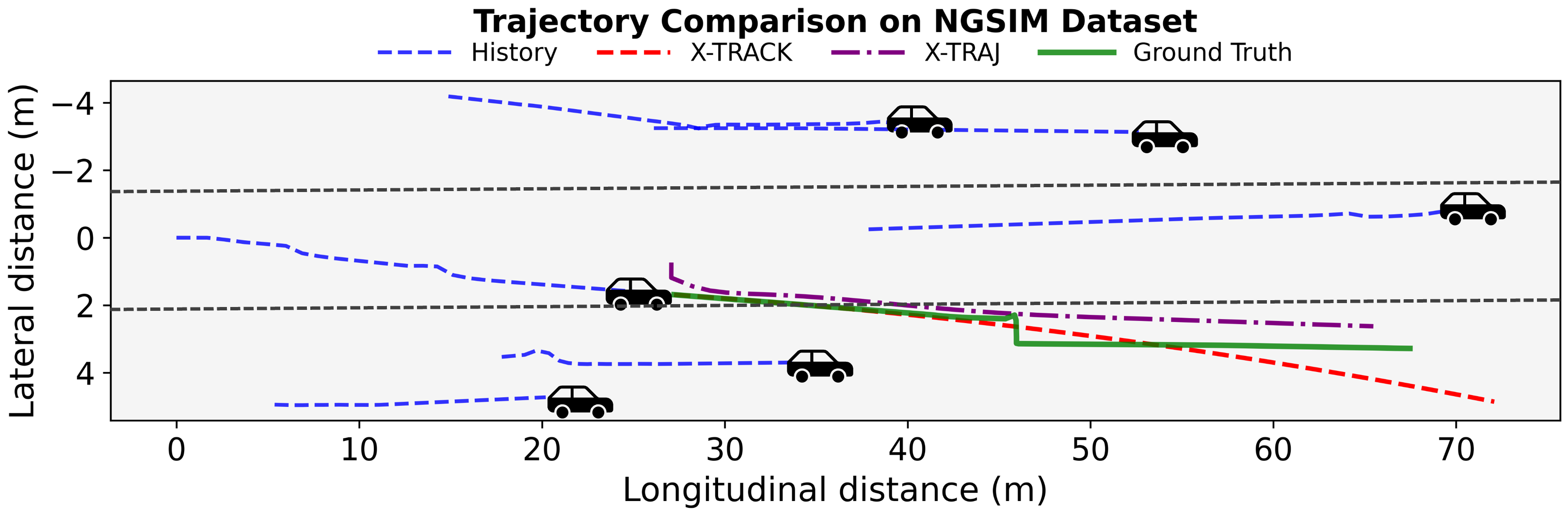}
    \caption{Non-realistic ground truth trajectory along with X-TRAJ and X-TRACK predictions on NGSIM.}
    \label{fig:ngsim}
\end{figure}

An extensive statistical analysis by Coifman et al. \cite{COIFMAN2017362} conclusively shows that the NGSIM dataset contains significant annotation inaccuracies, giving rise to complex and physically implausible vehicle motion patterns as seen in Fig \ref{fig:ngsim}. In contrast, highD consists of well-annotated highway scenarios with substantially smoother and more physically consistent trajectories. Notably, Coifman et al. \cite{COIFMAN2017362} observed that NGSIM acceleration often exhibits unrealistically large magnitudes, demonstrating that these inaccuracies disproportionately affect models that rely on acceleration either as an input or an output. As X-TRACK explicitly enforces constraints on acceleration through its kinematic layer, its performance is directly impacted by such inaccuracies. 

As a result, while X-TRAJ is able to overfit these annotation artifacts and achieve lower displacement errors, X-TRACK prioritizes physically consistent motion, which may lead to higher displacement errors when evaluated against noisy or physically implausible ground truth labels. This directly explains why the kinematic layer yields significant performance improvements on highD but does not consistently translate into gains on NGSIM.

Another contributing factor to the weaker performance on NGSIM is the limited number of training samples extracted to create a balanced dataset. Since the original NGSIM data is severely imbalanced, training on the entire dataset would bias the model toward dominant keep-lane behavior. Therefore, after preprocessing, only $2337$ scenarios were retained, approximately $6$ times fewer than in highD. This reduction further increases the sensitivity of physics-constrained models to annotation noise and limits the model's ability to generalize.

\subsection{Ablation Study on xLSTM Variants in X-TRACK}
\label{sec:ablation}
\begin{table}[htbp]
\caption{ADE and FDE metrics to compare different encoder-decoder combinations in X-TRACK on highD dataset}
\begin{center}
\renewcommand{\arraystretch}{1.2}
\begin{tabular}{|l|l|c|c|}
\hline
\textbf{Encoder} & \textbf{Decoder} & \textbf{ADE[m]} & \textbf{FDE[m]@5s} \\
\hline
LSTM & sLSTM   & 0.63 & 1.78 \\
sLSTM & sLSTM  & \underline{0.57} & 1.78 \\
LSTM & LSTM    & 0.65 & 1.85 \\
sLSTM & LSTM   & \textbf{0.56} & \textbf{1.76} \\
\hline
\end{tabular}
\label{tab:ablation_encdec}
\end{center}
\end{table}
Table \ref{tab:ablation_encdec} shows the ADE and FDE metrics of X-TRACK on the highD dataset, using either LSTM or sLSTM for the encoder-decoder combination. With LSTM as decoder, switching the encoder from LSTM to sLSTM gives better ADE and FDE metrics. Similarly, with sLSTM as decoder, switching the encoder from LSTM to sLSTM also produces lower errors. It was observed that an sLSTM encoder and an LSTM decoder yield the lowest error among all configurations. The worst performance is observed when LSTM is used both as an encoder and as a decoder, and all the other configurations show similar performance. The sLSTM encoder in combination with the LSTM decoder achieves $13.85\%$ and $4.86\%$ performance improvement in ADE and FDE on highD, respectively, than the conventional LSTM encoder-decoder architecture with an integrated physics-based kinematic model.
\section{Conclusion}
This paper introduced a novel xLSTM-based framework for highway vehicle trajectory prediction. The paper presents X-TRAJ as the first application of xLSTM to trajectory prediction, along with its physics-augmented extension X-TRACK. On highD, X-TRACK reduces the prediction error by $78.7\%$ and $19.7\%$ at $1$-second and $5$-second prediction, respectively, corresponding to approximately $4.7\times$ and $1.25\times$ lower error compared to the strongest baseline, GFTNNv2 \cite{GFTNNV2}.

In the case of highD, X-TRACK achieves an improvement of up to $79 \%$ at a $1$-second and about $32 \%$ at a $5$-second prediction horizon as compared to X-TRAJ, demonstrating that the use of non-holonomic vehicle constraints leads to smoother and physically consistent predictions. The kinematic module enables more accurate prediction of highway trajectories as the physical laws of vehicle dynamics are known and need not be inferred from the data in the case of deep learning-based models. 

X-TRACK consistently outperforms state-of-the-art baselines in the case of the highD dataset. The evaluation on the NGSIM dataset indicates that X-TRACK is among the state-of-the-art models, though with the annotation inaccuracies causing non-realistic driving trajectories and the reduced dataset size in NGSIM, making it difficult for X-TRACK to predict accurate trajectories, and therefore, the results are less statistically significant as compared to the highD dataset.

These findings suggest that physics-based priors, when combined with data-driven sequence models, bridge the gap between predictive accuracy and physical consistency for highway trajectory prediction. While our current scope is limited to highway scenarios, future work will extend the framework by integrating richer contextual information, such as detailed road structure and HD map data, to improve scenario representation. Incorporating multi-modal prediction and uncertainty quantification offers a natural extension to capture multiple plausible future trajectories. Furthermore, the proposed xLSTM architecture could also be extended to urban driving datasets in order to evaluate its performance under more diverse and complex traffic environments. 

\section*{Acknowledgment}
This research was supported by Technische Hochschule Augsburg and the Hightech Agenda Bavaria, funded by the Free State of Bavaria, Germany. The authors would like to thank their colleagues at Data Science und Autonome Systeme Technologietransferzentrum (TTZ) Landsberg for their insightful discussions and support.
\bibliographystyle{IEEEtran}
\bibliography{references}

\end{document}